\documentclass[12pt,twoside]{article}
\usepackage{latexsym,amsmath,amssymb,hyperref}
\def\,{\mskip 3mu} \def\>{\mskip 4mu plus 2mu minus 4mu} \def\;{\mskip 5mu plus 5mu} \def\!{\mskip-3mu}
\def\dispmuskip{\thinmuskip= 3mu plus 0mu minus 2mu \medmuskip=  4mu plus 2mu minus 2mu \thickmuskip=5mu plus 5mu minus 2mu}
\def\textmuskip{\thinmuskip= 0mu                    \medmuskip=  1mu plus 1mu minus 1mu \thickmuskip=2mu plus 3mu minus 1mu}
\textmuskip
\def\beq{\dispmuskip\begin{equation}}    \def\eeq{\end{equation}\textmuskip}
\def\beqn{\dispmuskip\begin{displaymath}}\def\eeqn{\end{displaymath}\textmuskip}
\def\bqa{\dispmuskip\begin{eqnarray}}    \def\eqa{\end{eqnarray}\textmuskip}
\def\bqan{\dispmuskip\begin{eqnarray*}}  \def\eqan{\end{eqnarray*}\textmuskip}

\newtheorem{theorem}{Theorem}

\newtheorem{lemma}[theorem]{Lemma}
\newtheorem{definition}[theorem]{Definition}
\newtheorem{myexample}[theorem]{Example}
\newtheorem{proposition}[theorem]{Proposition}
\newenvironment{keywords}{\centerline{\bf\small
Keywords}\begin{quote}\small}{\par\end{quote}\vskip 1ex}
\def\paradot#1{\vspace{1.3ex plus 0.5ex minus 0.5ex}\noindent{\bf\boldmath{#1.}}}

\def\aidx#1{}
\def\req#1{(\ref{#1})}
\def\toinfty#1{\smash{\stackrel{#1\to\infty}{\longrightarrow}}}
\def\iffs{\Leftrightarrow}
\def\eps{\varepsilon}
\def\nq{\hspace{-1em}}
\def\qed{\hspace*{\fill}\rule{1.4ex}{1.4ex}$\quad$\\}
\def\odt{{\textstyle{1\over 2}}}

\def\SetR{I\!\!R}
\def\SetN{I\!\!N}

\def\qmbox#1{{\quad\mbox{#1}\quad}}

\def\v{\vec}

\def\lb{{\log_2}}

\def\v{\boldsymbol} 
\def\text#1{\mbox{\scriptsize #1}}

\def\g{\gamma}
\def\G{\Gamma}
\def\d{\delta}
\def\a{\alpha}
\def\b{\beta}
\def\low{\underline}
\def\up{\overline}
\def\fract#1#2{{\textstyle{#1\over#2}}}
\def\eh{h^{e\!f\!f}}               
\def\ph{h^{\text{\it quasi}}}      

\begin{document}

\title{\vspace{-3ex}\normalsize\sc Technical Report \hfill IDSIA-11-06
\vskip 2mm\bf\Large\hrule height5pt \vskip 6mm
General Discounting versus Average Reward
\vskip 6mm \hrule height2pt}
\author{{\bf Marcus Hutter}\\[3mm]
\normalsize IDSIA, Galleria 2, CH-6928\ Manno-Lugano, Switzerland\\
\normalsize marcus@idsia.ch \hspace{9ex} http://www.idsia.ch/$^{_{_\sim}}\!$marcus }
\date{January 2006}
\maketitle

\begin{abstract}\noindent
Consider an agent interacting with an environment in cycles. In
every interaction cycle the agent is rewarded for its performance.
We compare the average reward $U$ from cycle $1$ to $m$ (average
value) with the future discounted reward $V$ from cycle $k$ to
$\infty$ (discounted value). We consider essentially arbitrary
(non-geometric) discount sequences and arbitrary reward sequences
(non-MDP environments). We show that asymptotically $U$ for
$m\to\infty$ and $V$ for $k\to\infty$ are equal, provided both
limits exist. Further, if the effective horizon grows linearly
with $k$ or faster, then the existence of the limit of $U$ implies
that the limit of $V$ exists. Conversely, if the effective horizon
grows linearly with $k$ or slower, then existence of the limit of
$V$ implies that the limit of $U$ exists.
\def\contentsname{\centering\normalsize Contents}
{\parskip=-2.5ex\tableofcontents}
\end{abstract}


\newpage
\section{Introduction}\label{secInt}

We consider the reinforcement learning setup
\cite{Russell:03,Hutter:04uaibook}, where an agent interacts with
an environment in cycles. In cycle $k$, the agent outputs (acts)
$a_k$, then it makes observation $o_k$ and receives reward $r_k$,
both provided by the environment. Then the next cycle $k+1$
starts. For simplicity we assume that agent and environment are
deterministic.

Typically one is interested in action sequences, called plans or
policies, for agents that result in high reward. The simplest
reasonable measure of performance is the total reward sum or
equivalently the average reward, called average value
$U_{1m}:={1\over m}[r_1+...+r_m]$, where $m$ should be the
lifespan of the agent. One problem is that the lifetime is often
not known in advance, e.g.\ often the time one is willing to let a
system run depends on its displayed performance. More serious is
that the measure is indifferent to whether an agent receives high
rewards early or late if the values are the same.

A natural (non-arbitrary) choice for $m$ is to consider the limit
$m\to\infty$. While the indifference may be acceptable for finite
$m$, it can be catastrophic for $m=\infty$. Consider an agent that
receives no reward until its first action is $b_k=b$, and then
once receives reward $k-1\over k$. For finite $m$, the optimal $k$
to switch from action $a$ to $b$ is $k_{opt}=m$. Hence
$k_{opt}\to\infty$ for $m\to\infty$, so the reward maximizing
agent for $m\to\infty$ actually always acts with $a$, and hence
has zero reward, although a value arbitrarily close to 1 would be
achievable. (Immortal agents are lazy \cite[Sec.5.7]{Hutter:04uaibook}).
More serious, in general the limit $U_{1\infty}$ may not even
exist.

Another approach is to consider a moving horizon. In cycle $k$,
the agent tries to maximize $U_{km}:={1\over m-k+1}[r_k+...+r_m]$,
where $m$ increases with $k$, e.g.\ $m=k+h-1$ with $h$ being the
horizon. This naive truncation is often used in games like chess
(plus a heuristic reward in cycle $m$) to get a reasonably small
search tree. While this can work in practice, it can lead to
inconsistent optimal strategies, i.e.\ to agents that change their
mind. Consider the example above with $h=2$. In every cycle $k$ it
is better first to act $a$ and then $b$
($U_{km}=r_k+r_{k+1}=0+{k\over k+1}$), rather than immediately $b$
($U_{km}=r_k+r_{k+1}={k-1\over k}+0$), or $a,a$ ($U_{km}=0+0$).
But entering the next cycle $k+1$, the agent throws its original plan
overboard, to now choose $a$ in favor of $b$, followed by $b$.
This pattern repeats, resulting in no reward at all.

The standard solution to the above problems is to consider
geometrically=exponentially discounted reward
\cite{Samuelson:37,Bertsekas:96,Sutton:98}. One discounts the reward for every
cycle of delay by a factor $\g<1$, i.e.\ considers
$V_{k\g} := (1-\g)\sum_{i=k}^\infty\g^{i-k} r_i$. The
$V_{1\g}$ maximizing policy is consistent in the sense that its
actions $a_k,a_{k+1},...$ coincide with the optimal policy based
on $V_{k\g}$. At first glance, there seems to be no arbitrary lifetime
$m$ or horizon $h$, but this is an illusion. $V_{k\g}$ is
dominated by contributions from rewards
$r_k...r_{k+O(\ln\g^{-1})}$, so has an effective horizon
$\eh\approx\ln\g^{-1}$. While such a sliding effective horizon
does not cause inconsistent policies, it can nevertheless
lead to suboptimal behavior. For every (effective) horizon, there
is a task that needs a larger horizon to be solved. For
instance, while $\eh=5$ is sufficient for tic-tac-toe, it is
definitely insufficient for chess. There are elegant closed form
solutions for Bandit problems, which show that for any $\g<1$, the
Bayes-optimal policy can get stuck with a suboptimal arm (is
not self-optimizing) \cite{Berry:85,Kumar:86}.

For $\g\to 1$, $\eh\to\infty$, and the defect decreases. There are
various deep papers considering the limit $\g\to 1$
\cite{Kelly:81}, and comparing it to the limit $m\to\infty$ \cite{Kakade:01}.
The analysis is typically restricted to ergodic MDPs for which the
limits $\lim_{\g\to 1}V_{1\g}$ and $\lim_{m\to\infty}U_{1m}$
exist. But like the limit policy for $m\to\infty$, the limit
policy for $\g\to 1$ can display very poor performance, i.e.\ we
need to choose $\g<1$ fixed in advance (but how?), or consider
higher order terms \cite{Mahadevan:96,Avrachenkov:99}. We also
cannot consistently adapt $\g$ with $k$. Finally, the value limits
may not exist beyond ergodic MDPs.

There is little work on other than geometric discounts. In the
psychology and economics literature it has been argued that people
discount a one day=cycle delay in reward more if it concerns
rewards now rather than later, e.g.\ in a year (plus one day)
\cite{Frederick:02}. So there is some work on ``sliding'' discount
sequences $W_{k\g}\propto \g_0 r_k+\g_1 r_{k+1}+...$. One can show
that this also leads to inconsistent policies if $\v\g$ is
non-geometric \cite{Strotz:55,Vieille:04}.

Is there any non-geometric discount leading to consistent
policies? In \cite{Hutter:02selfopt} the generally discounted
value $V_{k\g}:={1\over\G_k}\sum_{i=k}^\infty\g_i r_i$ with
$\G_k:=\sum_{i=k}^\infty\g_i<\infty$ has been introduced. It is
well-defined for arbitrary environments, leads to consistent
policies, and e.g.\ for quadratic discount $\g_k=1/k^2$ to an
increasing effective horizon (proportionally to $k$), i.e.\ the
optimal agent becomes increasingly farsighted in a consistent way,
leads to self-optimizing policies in ergodic ($k$th-order) MDPs in
general, Bandits in particular, and even beyond MDPs. See
\cite{Hutter:02selfopt} for these and \cite{Hutter:04uaibook} for
more results. The only other serious analysis of general discounts
we are aware of is in \cite{Berry:85}, but their analysis is
limited to Bandits and so-called regular discount. This discount
has bounded effective horizon, so also does not lead to
self-optimizing policies.

The {\em asymptotic} total average performance $U_{1\infty}$ and
future discounted performance $V_{\infty\g}$ are of key interest.
For instance, often we do not know the exact environment in
advance but have to {\em learn} it from past experience, which is
the domain of reinforcement learning \cite{Sutton:98} and adaptive
control theory \cite{Kumar:86}. Ideally we would like a learning
agent that performs {\em asymptotically} as well as the optimal
agent that knows the environment in advance.

\paradot{Contents and main results}
The subject of study of this paper is the relation between
$U_{1\infty}$ and $V_{\infty\g}$ for {\em general discount} $\v\g$
and {\em arbitrary environment}. The importance of the performance
measures $U$ and $V$, and general discount $\v\g$ has been
discussed above. There is also a clear need to study general
environments beyond ergodic MDPs, since the real world is neither
ergodic (e.g.\ losing an arm is irreversible) nor completely
observable.

The only restriction we impose on the discount sequence $\v\g$ is
summability ($\G_1<\infty$) so that $V_{k\g}$ exists, and
monotonicity ($\g_k\geq\g_{k+1}$).
Our main result is that if both limits $U_{1\infty}$ and
$V_{\infty\g}$ exist, then they are necessarily equal (Section
\ref{secAEDV}, Theorem \ref{thUeqV}). Somewhat surprisingly this
holds for {\em any} discount sequence $\v\g$ and {\em any}
environment (reward sequence $\v r$), whatsoever.

Note that limit $U_{1\infty}$ may exist or not, independent of
whether $V_{\infty\g}$ exists or not. We present examples of the
four possibilities in Section \ref{secEx}.
Under certain conditions on $\v\g$, existence of $U_{1\infty}$
implies existence of $V_{\infty\g}$, or vice versa. We show that
if (a quantity closely related to) the effective horizon grows
linearly with $k$ or faster, then existence of $U_{1\infty}$
implies existence of $V_{\infty\g}$ and their equality (Section
\ref{secAIDV}, Theorem \ref{thUimpV}). Conversely, if the
effective horizon grows linearly with $k$ or slower, then
existence of $V_{\infty\g}$ implies existence of $U_{1\infty}$ and
their equality (Section \ref{secDIAV}, Theorem \ref{thVimpU}).
Note that apart from discounts with oscillating effective horizons,
this implies (and this is actually the path used to prove) the
first mentioned main result.
In Sections \ref{secAV} and \ref{secDV} we define and provide some
basic properties of average and discounted value, respectively.

\section{Example Discount and Reward Sequences}\label{secEx}

In order to get a better feeling for general discount sequences,
effective horizons, average and discounted value, and their
relation and existence, we first consider various examples.

\paradot{Notation}
In the following we assume that $i,k,m,n\in\SetN$ are natural
numbers, $\low F:=\low\lim_n F_n=\lim_{k\to\infty}\inf_{n>k}F_n$ denotes the
limit inferior and $\up F:=\up\lim_n F_n=\lim_{k\to\infty}\sup_{n>k}F_n$ the
limit superior of $F_n$, $\forall'n$ means for all but finitely many $n$,
$\v\g=(\g_1,\g_2,...)$ denotes a summable discount sequence in the
sense that $\G_k:=\sum_{i=k}^\infty\g_i<\infty$ and
$\g_k\in\SetR^+$ $\forall k$, $\v r=(r_1,r_2,...)$ is a bounded
reward sequence w.l.g.\ $r_k\in[0,1]$ $\forall k$,
constants $\a,\b\in[0,1]$, boundaries $0\leq
k_1<m_1<k_2<m_2<k_3<...$, total average value $U_{1m}:={1\over
m}\sum_{i=1}^m r_i$ (see Definition \ref{defAvVal}) and future
discounted value $V_{k\g}={1\over\G_k}\sum_{i=k}^\infty\g_i r_i$
(see Definition \ref{defDiscVal}).
%
The derived theorems also apply to general bounded rewards $r_i\in[a,b]$
by linearly rescaling $r_i\leadsto {r_i-a\over b-a}\in[0,1]$ and
$U\leadsto {U-a\over b-a}$ and $V\leadsto {V-a\over b-a}$.

\paradot{Discount sequences and effective horizons}
Rewards $r_{k+h}$ give only a small contribution to $V_{k\g}$ for
large $h$, since $\g_{k+h}\toinfty{h}0$. More important, the whole
reward tail from $k+h$ to $\infty$ in $V_{k\g}$ is bounded by
${1\over\G_k}[\g_{k+h}+\g_{k+h+1}+...]$, which tends to zero for
$h\to\infty$. So effectively $V_{k\g}$ has a horizon $h$ for which
the cumulative tail weight $\G_{k+h}/\G_k$ is, say, about
$\odt$, or more formally $\eh_k:=\min\{h\geq
0:\G_{k+h}\leq\odt\G_k\}$.
The closely related quantity $\ph_k:=\G_k/\g_k$,
which we call the quasi-horizon, will play an important role in
this work.
The following table summarizes various discounts with their properties.
\beqn\arraycolsep4pt
\begin{array}{l||c|c|c|c|cl}
  \mbox{Discounts}
    & \g_k
    & \G_k
    & \eh_k
    & \ph_k
    & k\g_k/\G_k & \to ?
  \\ \hline\hline
  \mbox{finite}
    & {1 \;\mbox{\scriptsize for}\; k\leq m \atop 0 \;\mbox{\scriptsize for}\; k>m}
    & m-k+1
    & \odt(m-k+1)
    & m-k+1
    & {k\over m-k+1} &
  \\ \hline
  \mbox{geometric}
    & \g^k, \; 0\leq\g<1
    & {\g^k\over 1-\g}
    & {\ln 2\over\ln\g^{-1}}
    & {1\over 1-\g}
    & (1-\g)k & \to\infty
  \\ \hline
  \mbox{quadratic}
    & {1\over k(k+1)}
    & {1\over k}
    & k
    & k+1
    & {k\over k+1} & \to 1
  \\ \hline
  \mbox{power}
    & k^{-1-\eps}, \; \eps>0
    & \sim {1\over\eps}k^{-\eps}
    & \sim (2^{1/\eps}-1) k
    & \sim {k\over\eps}
    & \sim\eps & \to \eps
  \\ \hline
  \mbox{harmonic$_\approx$}
    & {1\over k\ln^2 k}
    & \sim{1\over\ln k}
    & \sim k^2
    & \sim k\ln k
    & \sim{1\over\ln k} & \to 0
\end{array}
\eeqn
\noindent For instance, the standard discount is geometric
$\g_k=\g^k$ for some $0\leq\g<1$, with constant effective horizon
${\ln(1/2)\over\ln\g}$. (An agent with $\g=0.95$ can/will not plan
farther than about 10-20 cycles ahead).
Since in this work we allow for general discount, we can even
recover the average value $U_{1m}$ by choosing $\g_k=\{ {1
\;\mbox{\scriptsize for}\; k\leq m\atop 0 \;\mbox{\scriptsize
for}\; k>m}\}$.
A power discount $\g_k=k^{-\alpha}$ ($\alpha>1$) is very
interesting, since it leads to a linearly increasing effective
horizon $\eh_k\propto k$, i.e.\ to an agent whose farsightedness
increases proportionally with age.
This choice has some appeal, as it avoids preselection of a global
time-scale like $m$ or ${1\over 1-\g}$, and it seems that humans
of age $k$ years usually do not plan their lives for more than,
perhaps, the next $k$ years. It is also the boundary case for
which $U_{1\infty}$ exists if and only if $V_{\infty\g}$ exists.

\paradot{Example reward sequences}
Most of our (counter)examples will be for binary reward $\v
r\in\{0,1\}^\infty$. We call a maximal consecutive subsequence of
ones a 1-run. We denote start, end, and length of the $n$th run
by $k_n$, $m_n-1$, and $A_n=m_n-k_n$, respectively. The following
0-run starts at $m_n$, ends at $k_{n+1}-1$, and has length
$B_n=k_{n+1}-m_n$. The (non-normalized) discount sum in 1/0-run
$n$ is denoted by $a_n$ / $b_n$, respectively.
The following definition and two lemmas facilitate the discussion
of our examples. The proofs contain further useful relations.

\begin{definition}[Value for binary rewards]\label{defBin}
Every binary reward sequence $\v r\in\{0,1\}^\infty$ can be defined by
the sequence of change points $0\leq k_1<m_1<k_2<m_2<...$ with
\beqn
  r_k=1 \quad\iff\quad k\in\bigcup_n{\cal S}_n, \qmbox{where}
  {\cal S}_n:=\{k\in\SetN: k_n\leq k<m_n\}.
\eeqn
\end{definition}

The intuition behind the following lemma is that the relative
length $A_n$ of a 1-run and the following 0-run $B_n$ (previous
0-run $B_{n-1}$) asymptotically provides a lower (upper) limit of
the average value $U_{1m}$.

\begin{lemma}[Average value for binary rewards]\label{lemAvBin}
For binary $\v r$ of Definition \ref{defBin}, let $A_n:=m_n-k_n$
and $B_n:=k_{n+1}-m_n$ be the lengths of the $n$th 1/0-run. Then
\bqan
  \mbox{If} & {A_n\over A_n+B_n}\to\a \qmbox{then} \low U_{1\infty}=\lim_n U_{1,k_n-1}=\a
\\
  \mbox{If} & {A_n\over B_{n-1}+A_n}\to\b \qmbox{then} \up U_{1\infty}=\lim_n U_{1,m_n-1}=\b
\eqan
In particular, if $\a=\b$, then $U_{1\infty}=\a=\b$ exists.
\end{lemma}

\paradot{Proof}
The elementary identity $U_{1m}=U_{1,m-1}+{1\over
m}(r_m-U_{1,m-1})\gtrless U_{1,m-1}$ if $r_m=\{ {1\atop 0} \}$ implies
\bqan
  U_{1k_n} \leq U_{1m} \leq U_{1,m_n-1} & \mbox{for} & k_n\leq m<m_n
\\
  U_{1,k_{n+1}-1} \leq U_{1m} \leq U_{1,m_n} & \mbox{for} & m_n\leq m<k_{n+1}
\eqan
\bqa\nonumber
  & \Rightarrow\quad & \displaystyle \inf_{n\geq n_0} U_{1k_n} \;\leq\; U_{1m}
                \;\leq\; \sup_{m\geq n_0} U_{1,m_n-1} \quad \forall m\geq k_{n_0}
\\ \label{prAvBin3}
  & \Rightarrow\quad & \displaystyle \mathop{\low\lim}_n U_{1k_n}
                \;=\; \low U_{1\infty} \;\leq\; \up U_{1\infty}
                \;=\; \mathop{\up\lim}_n U_{1,m_n-1}
\eqa
Note the equalities in the last line. The $\geq$ holds, since
$(U_{1k_n})$ and $(U_{1,m_n-1})$ are subsequences of $(U_{1m})$.
Now
\beq\label{prAvBin1}\textstyle
  \mbox{If}\quad {A_n\over A_n+B_n}\geq\a \;\forall n \qmbox{then}
  U_{1,k_n-1} = {A_1\;+\;...\;+\;A_{n-1}\over A_1+B_1+...+A_{n-1}+B_{n-1}}\geq\a \;\forall n
\eeq
This implies $\inf_n{A_n\over A_n+B_n}\leq\inf_n U_{1,k_n-1}$. If
the condition in \req{prAvBin1} is initially (for a finite number
of $n$) violated, the conclusion in \req{prAvBin1} still holds
asymptotically. A standard argument along these lines shows that
we can replace the $\inf$ by a $\low\lim$, i.e.\
\beqn
  \mathop{\low\lim}_n\fract{A_n}{A_n+B_n}\leq\mathop{\low\lim}_n U_{1,k_n-1}
  \qmbox{and similarly}
  \mathop{\up\lim}_n\fract{A_n}{A_n+B_n}\geq\mathop{\up\lim}_n U_{1,k_n-1}
\eeqn
Together this shows that $\lim_n U_{1,k_n-1}=\a$ exists, if
$\lim_n{A_n\over A_n+B_n}=\a$ exists.
Similarly
\beq\label{prAvBin2}\textstyle
  \mbox{If}\quad {A_n\over B_{n-1}+A_n}\geq\b \;\forall n \qmbox{then}
  U_{1,m_n-1} = {A_1\;+\;...\;+\;A_n\over B_0+A_1+...+B_{n-1}+A_n}\geq\b \;\forall n
\eeq
where $B_0:=0$.
This implies $\inf_n{A_n\over B_{n-1}+A_n}\leq\inf_n U_{1,m_n-1}$,
and by an asymptotic refinement of \req{prAvBin2}
\beqn
  \mathop{\low\lim}_n\fract{A_n}{B_{n-1}+A_n}\leq\mathop{\low\lim}_n U_{1,m_n-1}
  \qmbox{and similarly}
  \mathop{\up\lim}_n\fract{A_n}{B_{n-1}+A_n}\geq\mathop{\up\lim}_n U_{1,m_n-1}
\eeqn
Together this shows that $\lim_n U_{1,m_n-1}=\b$ exists, if
$\lim_n{A_n\over B_{n-1}+A_n}=\b$ exists.
\qed

Similarly to Lemma \ref{lemAvBin}, the asymptotic ratio of the
discounted value $a_n$ of a 1-run and the discount sum $b_n$ of
the following ($b_{n-1}$ of the previous) 0-run determines
the upper (lower) limits of the discounted value $V_{k\g}$.

\begin{lemma}[Discounted value for binary rewards]\label{lemDiscBin}
For binary $\v r$ of Definition \ref{defBin}, let
$a_n:=\sum_{i=k_n}^{m_n-1}\g_i=\G_{k_n}-\G_{m_n}$ and
$b_n:=\sum_{i=m_n}^{k_{n+1}-1}\g_i=\G_{m_n}-\G_{k_{n+1}}$
be the discount sums of the $n$th 1/0-run. Then
\bqan
  \mbox{If} & {a_{n+1}\over b_n+a_{n+1}}\to\a \qmbox{then} \low V_{\infty\g}=\lim_n V_{m_n\g}=\a
\\
  \mbox{If} & {a_n\over a_n+b_n}\to\b \qmbox{then} \up V_{\infty\g}=\lim_n V_{k_n\g}=\b
\eqan
In particular, if $\a=\b$, then $V_{\infty\g}=\a=\b$ exists.
\end{lemma}

\paradot{Proof} The proof is very similar to the proof of Lemma \ref{lemAvBin}.
The elementary identity
$V_{k\g}=V_{k+1,\g}+{\g_k\over\G_k}(r_k-V_{k+1,\g})\gtrless
V_{k+1,\g}$ if $r_k=\{ {1\atop 0} \}$ implies
\bqan
  V_{m_n\g} \leq V_{k\g} \leq V_{k_n\g} & \mbox{for} & k_n\leq k\leq m_n
\\
  V_{m_n\g} \leq V_{k\g} \leq V_{k_{n+1}\g} & \mbox{for} & m_n\leq k\leq k_{n+1}
\eqan
\bqa\nonumber
  & \Rightarrow\quad & \displaystyle \inf_{n\geq n_0} V_{m_n\g} \;\leq\; V_{k\g}
                \;\leq\; \sup_{m\geq n_0} V_{k_n\g} \quad \forall k\geq k_{n_0}
\\ \label{prDiscBin1}
  & \Rightarrow\quad & \displaystyle \mathop{\low\lim}_n V_{m_n\g}
                \;=\; \low V_{\infty\g} \;\leq\; \up V_{\infty\g}
                \;=\; \mathop{\up\lim}_n V_{k_n\g}
\eqa
Note the equalities in the last line. The $\geq$ holds, since
$(V_{k_n\g})$ and $(V_{m_n\g})$ are subsequences of $(V_{k\g})$.
Now if ${a_n\over a_n+b_n}\geq\b \;\forall n\geq n_0$ then
$V_{k_n\g} = {a_n\;+\;a_{n+1}\;+\;...\over
a_n+b_n+a_{n+1}+b_{n+1}+...}\geq\b$ $\forall n\geq n_0$. This
implies
\beqn
  \mathop{\low\lim}_n\fract{a_n}{a_n+b_n}\leq\mathop{\low\lim}_n V_{k_n\g}
  \qmbox{and similarly}
  \mathop{\up\lim}_n\fract{a_n}{a_n+b_n}\geq\mathop{\up\lim}_n V_{k_n\g}
\eeqn
Together this shows that $\lim_n V_{k_n\g}=\b$ exists, if
$\lim_n{a_n\over a_n+b_n}=\b$ exists.
Similarly if ${a_{n+1}\over b_n+a_{n+1}}\geq\a$ $\forall n\geq
n_0$ then $V_{m_n\g} = {a_{n+1}\;+\;a_{n+2}\;+...\over
b_n+a_{n+1}+b_{n+1}+a_{n+2}+...}\geq\a$ $\forall n\geq n_0$. This
implies
\beqn
  \mathop{\low\lim}_n\fract{a_{n+1}}{b_n+a_{n+1}}\leq\mathop{\low\lim}_n V_{m_n\g}
  \qmbox{and similarly}
  \mathop{\up\lim}_n\fract{a_{n+1}}{b_n+a_{n+1}}\geq\mathop{\up\lim}_n V_{m_n\g}
\eeqn
Together this shows that $\lim_n V_{m_n\g}=\a$ exists, if
$\lim_n{a_{n+1}\over b_n+a_{n+1}}=\a$ exists.
\qed

\begin{myexample}[\bf\boldmath $U_{1\infty}=V_{\infty\g}$]\rm\label{exUeqV}
Constant rewards $r_k\equiv\a$ is a trivial example for which
$U_{1\infty}=V_{\infty\g}=\a$ exist and are equal.

A more interesting example is $\v r=1^1 0^2 1^3
0^4...$ of linearly increasing 0/1-run-length with $A_n=2n-1$ and
$B_n=2n$, for which $U_{1\infty}=\odt$ exists. For quadratic
discount $\g_k={1\over k(k+1)}$, using $\G_k={1\over k}$,
$\ph_k=k+1=\Theta(k)$, $k_n=(2n-1)(n-1)+1$, $m_n=(2n-1)n+1$,
$a_n=\G_{k_n}-\G_{m_n}={A_n\over k_n m_n}\sim{1\over 2n^3}$, and
$b_n=\G_{m_n}-\G_{k_{n+1}}={B_n\over m_n k_{n+1}}\sim{1\over
2n^3}$, we also get $V_{\infty\g}=\odt$. The values converge, since
they average over increasingly many 1/0-runs, each of decreasing weight.
\end{myexample}

\begin{myexample}[\bf\boldmath simple $U_{1\infty}\not\Rightarrow V_{\infty\g}$]\rm\label{exUnVs}
Let us consider a very simple example with alternating rewards $\v
r=101010...$ and geometric discount $\g_k=\g^k$. It is immediate
that $U_{1\infty}=\odt$ exists, but $\low
V_{\infty\g}=V_{2k,\g}={\g\over 1+\g} < {1\over
1+\g}=V_{2k-1,\g}=\up V_{\infty\g}$.
\end{myexample}

\begin{myexample}[\bf\boldmath $U_{1\infty}\not\Rightarrow V_{\infty\g}$]\rm\label{exUnV}
Let us reconsider the more interesting example $\v r=1^1 0^2 1^3
0^4...$ of linearly increasing 0/1-run-length with $A_n=2n-1$ and
$B_n=2n$ for which $U_{1\infty}=\odt$ exists, as expected. On the
other hand, for geometric discount $\g_k=\g^k$, using
$\G_k={\g^k\over 1-\g}$ and
$a_n=\G_{k_n}-\G_{m_n}={\g^{k_n}\over 1-\g}[1-\g^{A_n}]$ and
$b_n=\G_{m_n}-\G_{k_{n+1}}={\g^{m_n}\over 1-\g}[1-\g^{B_n}]$,
i.e.\ ${b_n\over a_n}\sim\g^{A_n}\to 0$ and ${a_{n+1}\over
b_n}\sim\g^{B_n}\to 0$, we get $\low V_{\infty\g}=\a=0<1=\b=\up
V_{\infty\g}$. Again, this is plausible since for $k$ at the
beginning of a long run, $V_{k\g}$ is dominated by the reward 0/1
in this run, due to the bounded effective horizon of geometric $\v\g$.
\end{myexample}

\begin{myexample}[\bf\boldmath $V_{\infty\g}\not\Rightarrow U_{1\infty}$]\rm\label{exVnU}
Discounted may not imply average value on sequences of
exponentially increasing run-length like $\v
r=1^10^21^40^81^{16}...$ with $A_n=2^{2n-2}=k_n$ and
$B_n=2^{2n-1}=m_n$ for which $\low U_{1\infty}={A_n\over
A_n+B_n}={1\over 3}<{2\over 3}={A_n\over B_{n-1}+A_n}=\up
U_{1\infty}$, i.e.\ $U_{1\infty}$ does not exist.
On the other hand, $V_{\infty\g}$ exists for a discount with
super-linear horizon like $\g_k=[k\ln^2 k]^{-1}$, since
an increasing number of runs contribute to $V_{k\g}$:
$\G_k\sim{1\over\ln k}$, hence $\G_{k_n}\sim{1\over (2n-2)\ln 2}$
and $\G_{m_n}\sim{1\over (2n-1)\ln 2}$, which implies
$a_n=\G_{k_n}-\G_{m_n} \sim [4n^2\ln 2]^{-1} \sim \G_{m_n}-\G_{k_{n+1}}=b_n$,
i.e.\ $V_{\infty\g}=\odt$ exists.
\end{myexample}

\begin{myexample}[\bf\boldmath Non-monotone discount $\g$, $U_{1\infty}\neq V_{\infty\g}$]\rm\label{exUneqV}
Monotonicity of $\v\g$ in Theorems \ref{thUimpV}, \ref{thVimpU}, and
\ref{thUeqV} is necessary. As a simple counter-example consider
alternating rewards $r_{2k}=0$ with arbitrary $\g_{2k}$ and
$r_{2k-1}=1$ with $\g_{2k-1}=0$, which implies $V_{k\gamma}\equiv
0$, but $U_{1\infty}=\odt$.

The above counter-example is rather simplistic. One may hope
equivalence to hold on smoother $\v\g$ like
${\g_{k+1}\over\g_k}\to 1$. The following example shows that this
condition alone is not sufficient. For a counter-example one needs an
oscillating $\v\g$ of constant relative amplitude, but increasing
wavelength, e.g.\ $\g_k=[2+\cos(\pi\sqrt{2k})]/k^2$. For the
sequence $\v r=1^1 0^2 1^3 0^4...$ of Example \ref{exUnV} we had
$U_{1\infty}=\odt$. Using $m_n=\odt(2n-\odt)^2+{7\over 8}$ and
$k_{n+1}=\odt(2n+\odt)^2+{7\over 8}$, and
replacing the sums in the definitions of $a_n$ and $b_n$ by integrals,
we get $a_n\sim {1\over n^3}[\odt-{1\over\pi}]$ and
$b_n\sim {1\over n^3}[\odt+{1\over\pi}]$, which implies that
$V_{\infty\g}=\odt-{1\over\pi}$ exists, but differs from
$U_{1\infty}=\odt$.
\end{myexample}

\begin{myexample}[Oscillating horizon]\rm\label{exOsc}
It is easy to construct a discount $\v\g$ for which
$\sup_k{\G_k\over k\g_k}=\infty$ {\em and} $\sup_k{k\g_k\over \G_k}=\infty$
by alternatingly patching together discounts with super- and
sub-linear quasi-horizon $\ph_k$. For instance choose
$\g_k\propto\g^k$ geometric until ${\G_k\over k\g_k}<{1\over n}$,
then $\g_k\propto{1\over k\ln^2k}$ harmonic until ${\G_k\over
k\g_k}>n$, then repeat with $n\leadsto n+1$. The proportionality
constants can be chosen to insure monotonicity of $\v\g$. For
such $\v\g$ neither Theorem \ref{thUimpV} nor Theorem
\ref{thVimpU} is applicable, only Theorem \ref{thUeqV}.
\end{myexample}

\section{Average Value}\label{secAV}

We now take a closer look at the (total) average value $U_{1m}$ and
relate it to the future average value $U_{km}$, an intermediate
quantity we need later. We recall the definition of the average value:

\begin{definition}[Average value, \boldmath $U_{1m}$]\label{defAvVal}
Let $r_i\in[0,1]$ be the reward at time $i\in\SetN$. Then
\beqn
  U_{1m} \;:=\; {1\over m}\sum_{i=1}^m r_i \;\in[0,1]
\eeqn
is the average value
from time 1 to $m$, and $U_{1\infty}:=\lim_{m\to\infty}U_{1m}$ the
average value if it exists.
\end{definition}

We also need the average value $U_{km}:={1\over m-k+1}\sum_{i=k}^m
r_i$ from $k$ to $m$ and the following Lemma.

\begin{lemma}[Convergence of future average value, \boldmath $U_{k\infty}$]\label{lemUkm}
For $k_m\leq m\to\infty$ and every $k$ we have
\beqn
  U_{1m}\to\a \quad\iffs\quad U_{km}\to\a \quad\left.
  {\Rightarrow\quad U_{k_m m}\to\a  \qmbox{if}         \smash{\sup\limits_m}{k_m-1\over m}<1 \atop
   \Leftarrow\quad  U_{k_m m}\to\a \phantom{\qmbox{if} \smash{\sup\limits_m}{k_m-1\over m}<1} } \right.
\eeqn
\end{lemma}

The first equivalence states the obvious fact (and problem) that
any finite initial part has no influence on the average value
$U_{1\infty}$. Chunking together many $U_{k_m m}$ implies the last
$\Leftarrow$. The $\Rightarrow$ only works if we average in
$U_{k_m m}$ over sufficiently many rewards, which the stated
condition ensures ($\v r=101010...$ and $k_m=m$ is a simple
counter-example). Note that $U_{km_k}\to\a$ for $m_k\geq k\to\infty$
implies $U_{1m_k}\to\a$, but not necessarily $U_{1m}\to\a$ (e.g.\
in Example \ref{exVnU}, $U_{1m_k}={1\over 3}$ and ${k-1\over
m_k}\to 0$ imply $U_{km_k}\to{1\over 3}$ by \req{prUkm}, but
$U_{1\infty}$ does not exist).

\paradot{Proof}
The trivial identity $mU_{1m}=(k-1)U_{1,k-1}+(m-k+1)U_{km}$ implies
$U_{km}-U_{1m}={k-1\over m-k+1}(U_{1m}-U_{1,k-1})$ implies
\beq\label{prUkm}
  |U_{km}-U_{1m}|
  \;\leq\; {|U_{1m}-U_{1,k-1}|\over{m\over k-1}-1}
\eeq

$\iffs$) The numerator is bounded by 1, and for fixed $k$ and
$m\to\infty$ the denominator tends to $\infty$, which proves
$\iffs$.

$\Rightarrow$) We choose (small) $\eps>0$, $m_\eps$ large enough
so that $|U_{1m}-\a|<\eps$ $\forall m\geq m_\eps$, and
$m\geq{m_\eps\over\eps}$. If $k:=k_m\leq m_\eps$, then \req{prUkm}
is bounded by ${1\over 1/\eps-1}$. If $k:=k_m>m_\eps$, then
\req{prUkm} is bounded by ${2\eps\over 1/c-1}$, where
$c:=\sup_k{k_m-1\over m}<1$. This shows that $|U_{k_m
m}-U_{1m}|=O(\eps)$ for large $m$, which implies $U_{k_m m}\to\a$.

$\Leftarrow$) We partition the time-range
$\{1...m\}=\bigcup_{n=1}^L\{k_{m_n}...m_n\}$, where $m_1:=m$ and
$m_{n+1}:=k_{m_n}-1$. We choose (small) $\eps>0$, $m_\eps$ large
enough so that $|U_{k_m m}-\a|<\eps$ $\forall m\geq m_\eps$,
$m\geq{m_\eps\over\eps}$, and $l$ so that $k_{m_l}\leq m_\eps\leq
m_l$. Then
\bqan
  U_{1m} & = & {1\over m} \left[\sum_{n=1}^l+\sum_{n=l+1}^L\right] (m_n\!-\!k_{m_n}\!+\!1) U_{k_{m_n}m_n}
\\
  & \leq & {1\over m}\sum_{n=1}^l (m_n\!-\!k_{m_n}\!+\!1)(\a+\eps) + {m_{l+1}\!-\!k_{m_L}\!+\!1\over m}
\\
  & \leq & {m_1\!-\!k_{m_l}\!+\!1\over m}(\a+\eps) + {k_{m_l}\over m}
   \;\leq\; (\a+\eps) + \eps
\\
  \nq\nq \mbox{Similarly}\quad U_{1m} & \geq & {m_1\!-\!k_{m_l}\!+\!1\over m}(\a-\eps)
  \geq {m\!-\!m_\eps\over m}(\a-\eps) \geq (1-\eps)(\a-\eps)
\eqan
This shows that $|U_{1m}-\a|\leq 2\eps$ for sufficiently large
$m$, hence $U_{1m}\to\a$.
\qed

\section{Discounted Value}\label{secDV}

We now take a closer look at the (future) discounted value
$V_{k\g}$ for general discounts $\v\g$, and prove some useful
elementary asymptotic properties of discount $\g_k$ and normalizer
$\G_k$. We recall the definition of the discounted value:

\begin{definition}[Discounted value, \boldmath $V_{k\g}$]\label{defDiscVal}
Let $r_i\in[0,1]$ be the reward and $\g_i\geq 0$ a discount at
time $i\in\SetN$, where $\v\g$ is assumed to be summable in the
sense that $0<\G_k:=\sum_{i=k}^\infty\g_i<\infty$. Then
\beqn
  V_{k\g} \;:=\; {1\over\G_k}\sum_{i=k}^\infty\g_i r_i \;\in[0,1]
\eeqn
is the $\v\g$-discounted future value and
$V_{\infty\g}:=\lim_{k\to\infty}V_{k\g}$ its limit if it exists.
\end{definition}

We say that $\v\g$ is {\em monotone} if $\g_{k+1}\leq\g_k\forall
k$. Note that monotonicity and $\G_k>0$ $\forall k$ implies
$\g_k>0$ $\forall k$ and convexity of $\G_k$.

\begin{lemma}[Discount properties, \boldmath $\g/\G$]\label{lemDiscProp}
\bqan
  & i) & {\g_{k+1}\over\g_k} \to 1 \quad\Leftrightarrow\quad
         {\g_{k+\Delta}\over\g_k} \to 1 \quad\forall\Delta\in\SetN
\\
  & ii) & {\g_k\over\G_k}\to 0 \quad\Leftrightarrow\quad
          {\G_{k+1}\over\G_k}\to 1 \quad\Leftrightarrow\quad
         {\G_{k+\Delta}\over\G_k} \to 1 \quad\forall\Delta\in\SetN
\eqan
Furthermore, $(i)$ implies $(ii)$, but not necessarily the other way around
(even not if $\v\g$ is monotone).
\end{lemma}

\paradot{Proof}
$(i)\Rightarrow$
${\g_{k+\Delta}\over\g_k}=\prod_{i=k}^{\Delta-1}{\g_{i+1}\over\g_i} \toinfty{k} 1$, since
$\Delta$ is finite.
\\
$(i)\Leftarrow$ Set $\Delta=1$.
\\
$(ii)$ The first equivalence follows from $\G_k=\g_k+\G_{k+1}$.
The proof for the second equivalence is the same as for $(i)$ with
$\g$ replaced by $\G$.
\\
$(i)\Rightarrow(ii)$ Choose $\eps>0$. $(i)$ implies
${\g_{k+1}\over\g_k}\geq 1-\eps$ $\forall\,'k$ implies
\beqn
  \G_k \;=\; \sum_{i=k}^\infty\g_i
  \;=\; \g_k\sum_{i=k}^\infty \prod_{j=k}^{i-1}{\g_{i+1}\over\g_i}
  \;\geq\; \g_k\sum_{i=k}^\infty (1-\eps)^{i-k}
  \;=\; \g_k/\eps
\eeqn
hence ${\g_k\over\G_k}\leq\eps$ $\forall'k$, which implies
${\g_k\over\G_k}\to 0$.
\\
$(i)\not\Leftarrow(ii)$ Consider counter-example
$\g_k \;=\; 4^{-\lceil\lb k\rceil}$, i.e.
$\g_k=4^{-n}$ for $2^{n-1}<k\leq 2^n$.
Since $\G_k\geq\sum_{i=2^n}^\infty\g_i=2^{-n-1}$ we have
$0\leq{\g_k\over\G_k}\leq 2^{1-n}\to 0$, but
${\g_{k+1}\over\g_k}={1\over 4}\not\to 1$ for $k=2^n$.
\qed

\section{Average Implies Discounted Value}\label{secAIDV}

We now show that existence of $\lim_m U_{1m}$ can imply existence of
$\lim_k V_{k\g}$ and their equality. The necessary and sufficient
condition for this implication to hold is roughly that the
effective horizon grows linearly with $k$ or faster.
The auxiliary quantity $U_{km}$ is in a sense closer to $V_{k\g}$
than $U_{1m}$ is, since the former two both average from $k$
(approximately) to some (effective) horizon. If $\v\g$ is
sufficiently smooth, we can chop the area under the graph of
$V_{k\g}$ (as a function of $k$) ``vertically'' approximately into
a sum of average values, which implies

\begin{proposition}[Future average implies discounted value, \boldmath $U_\infty\Rightarrow V_{\infty\g}$]\label{propUimpV}
Assume $k\leq m_k\to\infty$ and monotone $\v\g$ with ${\g_{m_k}\over\g_k}\to 1$. If
$U_{km_k}\to\a$, then $V_{k\g}\to\a$.
\end{proposition}

The proof idea is as follows: Let $k_1=k$ and $k_{n+1}=m_{k_n}+1$.
Then for large $k$ we get
\bqan
  V_{k\g} & = & {1\over\G_k}\sum_{n=1}^\infty \sum_{i=k_n}^{m_{k_n}}\g_i r_i
  \;\approx\; {1\over\G_k}\sum_{n=1}^\infty \g_{k_n}(k_{n+1}-k_n)U_{k_n m_{k_n}}
\\
  & \approx & {\a\over\G_k}\sum_{n=1}^\infty \g_{k_n}(k_{n+1}-k_n)
  \;\approx\; {\a\over\G_k}\sum_{n=1}^\infty\sum_{i=k_n}^{m_{k_n}}\g_i
  \;=\; \a
\eqan
The (omitted) formal proof specifies the approximation error, which vanishes
for $k\to\infty$.

Actually we are more interested in relating the
(total) average value $U_{1\infty}$ to the (future) discounted
value $V_{k\g}$. The following (first main) Theorem shows that for
linearly or faster increasing quasi-horizon, we have
$V_{\infty\g}=U_{1\infty}$, provided the latter exists.

\begin{theorem}[Average implies discounted value, \boldmath $U_{1\infty}\Rightarrow V_{\infty\g}$]\label{thUimpV}\hfill\par
Assume $\sup_k{k\g_k\over\G_k}<\infty$ and monotone $\v\g$.
If $U_{1m}\to\a$, then $V_{k\g}\to\a$.
\end{theorem}

For instance, quadratic, power and harmonic discounts
satisfy the condition, but faster-than-power discount like geometric do not.
Note that Theorem \ref{thUimpV} does not imply Proposition \ref{propUimpV}.

The intuition of Theorem \ref{thUimpV} for binary reward is as
follows: For $U_{1m}$ being able to converge, the length of a run
must be small compared to the total length $m$ up to this run,
i.e.\ $o(m)$. The condition in Theorem \ref{thUimpV} ensures that
the quasi-horizon $\ph_k=\Omega(k)$ increases faster than the
run-lengths $o(k)$, hence $V_{k\g} \approx U_{k\Omega(k)} \approx
U_{1m}$ (Lemma \ref{lemUkm}) asymptotically averages over many
runs, hence should also exist. The formal proof ``horizontally''
slices $V_{k\g}$ into a weighted sum of average rewards $U_{1m}$.
Then $U_{1m}\to\a$ implies $V_{k\g}\to\a$.

\paradot{Proof}
We represent $V_{k\g}$ as a $\d_j$-weighted mixture of $U_{1j}$'s
for $j\geq k$, where $\d_j:=\g_j-\g_{j+1}\geq 0$. The condition
$\infty>c\geq{k\g_k\over\G_k}=:c_k$ ensures that the excessive
initial part $\propto U_{1,k-1}$ is ``negligible''. It is easy to
show that
\beqn
  \sum_{j=i}^\infty\d_j \;=\; \g_i \qmbox{and}
  \sum_{j=k}^\infty j\d_j \;=\; (k\!-\!1)\g_k+\G_k
\eeqn
We choose some (small) $\eps>0$, and $m_\eps$ large enough so that
$|U_{1m}-\a|<\eps$ $\forall m\geq m_\eps$. Then, for $k>m_\eps$ we get
\bqan
  V_{k\g} & = & {1\over\G_k}\sum_{i=k}^\infty\g_i r_i
  \;=\; {1\over\G_k}\sum_{i=k}^\infty\sum_{j=i}^\infty \d_j r_i
  \;=\; {1\over\G_k}\sum_{j=k}^\infty \sum_{i=k}^j \d_j r_i
\\
  & = & {1\over\G_k}\sum_{j=k}^\infty \d_j[jU_{1j}-(k\!-\!1)U_{1,k-1}]
\\
  & \lessgtr & {1\over\G_k}\sum_{j=k}^\infty \d_j[j(\a\pm\eps)-(k\!-\!1)(\a\mp\eps)]
\\
  & = & {1\over\G_k}[(k\!-\!1)\g_k+\G_k](\a\pm\eps)-{1\over\G_k}\g_k(k\!-\!1)(\a\mp\eps)
\\
  & = & \a \pm \Big( 1+{2(k-1)\g_k\over\G_k}\Big)\eps
  \;\lessgtr\; \a \pm (1+2c_k)\eps
\eqan
i.e.\ $|V_{k\g}-\a|<(1+2c_k)\eps\leq (1+2c)\eps$ $\forall k>m_\eps$, which implies
$V_{k\g}\to\a$.
\qed

Theorem \ref{thUimpV} can, for instance, be applied to Example \ref{exUeqV}.
Examples \ref{exUnVs}, \ref{exUnV}, and \ref{exUneqV} demonstrate
that the conditions in Theorem \ref{thUimpV} cannot be dropped.
The following proposition shows more strongly, that the sufficient
condition is actually necessary (modulo monotonicity of $\v\g$),
i.e.\ cannot be weakened.

\begin{proposition}[\bf\boldmath $U_{1\infty}\not\Rightarrow V_{\infty\g}$]\label{propUnV}\hfill\par
For every monotone $\v\g$ with $\sup_k{k\g_k\over\G_k}=\infty$, there
are $\v r$ for which $U_{1\infty}$
exists, but not $V_{\infty\g}$.
\end{proposition}

The proof idea is to construct a binary $\v r$ such that all change points
$k_n$ and $m_n$ satisfy $\G_{k_n}\approx 2\G_{m_n}$. This ensures
that $V_{k_n\g}$ receives a significant contribution from 1-run $n$,
i.e.\ is large. Choosing $k_{n+1}\gg m_n$ ensures that $V_{m_n\g}$ is
small, hence $V_{k\g}$ oscillates. Since the quasi-horizon
$\ph_k\neq\Omega(k)$ is small, the 1-runs are short
enough to keep $U_{1m}$ small so that $U_{1\infty}=0$.

\paradot{Proof}
The assumption ensures that there exists a sequence $m_1$, $m_2$, $m_3$, ...
for which
\beqn
  {m_n\g_{m_n}\over\G_{m_n}} \geq n^2
  \quad\mbox{We further (can) require $\G_{m_n}<\odt\G_{m_{n-1}+1}$
  $\quad(m_0:=0)$}
\eeqn
For each $m_n$ we choose $k_n$ such that $\G_{k_n}\approx
2\G_{m_n}$. More precisely, since $\G$ is monotone decreasing and
$\G_{m_n}<2\G_{m_n}\leq\G_{m_{n-1}+1}$, there exists (a unique)
$k_n$ in the range $m_{n-1}<k_n<m_n$ such that $\G_{{k_n}+1}<2\G_{m_n}\leq
\G_{k_n}$.
We choose a binary reward sequence with $r_k=1$ iff $k_n\leq
k<m_n$ for some $n$.
This implies
\bqan
  n^2 & \leq & {m_n\g_{m_n}\over\G_{m_n}}
  \;=\; {m_n\over m_n-k_n-1}{(m_n-k_n-1)\g_{m_n}\over\G_{m_n}}
\\
  & \leq & {m_n\over m_n-k_n-1}{\G_{k_n+1}-\G_{m_n}\over\G_{m_n}}
  \;\leq\;{m_n\over m_n-k_n-1}
\\
  \Longrightarrow\quad {m_n-k_n\over m_n} & = & {m_n-k_n-1\over m_n} +{1\over m_n}
  \;\leq\; {1\over n^2} + {\g_{m_n}\over\G_{m_n}}{1\over n^2}
  \;\leq\; {2\over n^2}
\\
  \Longrightarrow\quad U_{1m_n} & \leq & {1\over m_n}[k_l-1]+{1\over m_n}\sum_{n'=l}^n[m_{n'}-k_{n'}]
  \;\leq\; {k_l\over m_n}+\sum_{n'=l}^n{m_{n'}-k_{n'}\over m_{n'}}
\\
  & \leq & {k_l\over m_n}+\sum_{n'=l}^n{2\over {n'}^2}
  \;\leq\; {k_l\over m_n}+{2\over l-1}
\eqan
hence by \req{prAvBin3} we have $\up U_{1\infty} = \up\lim_n U_{1,m_n-1}
\;\leq {2\over l-1}$ $\forall l$, hence $U_{1\infty}=0$.
On the other hand
\beqn
  \G_{k_n}V_{k_n\g} \;=\; [\G_{k_n}\!-\!\G_{m_n}] + \G_{m_n}V_{m_n\g}
  \quad\Rightarrow\quad
  {1-V_{k_n\g}\over 1-V_{m_n\g}} \;=\; {\G_{m_n}\over\G_{k_n}} \;\leq\; \odt
\eeqn
This shows that $V_{k\g}$ cannot converge to an $\a<1$.
Theorem \ref{thUeqV} and $U_{1\infty}=0$ implies that $V_{k\g}$ can also
not converge to 1, hence $V_{\infty\g}$ does not exist.
\qed

\section{Discounted Implies Average Value}\label{secDIAV}

We now turn to the converse direction that existence of
$V_{\infty\g}$ can imply existence of $U_{1\infty}$ and their
equality, which holds under a nearly converse condition on the
discount: Roughly, the effective horizon has to grow linearly with
$k$ or slower.

\begin{theorem}[Discounted implies average value, \boldmath $V_{\infty\g}\Rightarrow U_{1\infty}$]\label{thVimpU}\hfill\par
Assume $\sup_k{\G_k\over k\g_k}<\infty$ and monotone $\v\g$.
If $V_{k\g}\to\a$, then $U_{1m}\to\a$.
\end{theorem}

For instance, power or faster and geometric discounts satisfy the
condition, but harmonic does not. Note that power discounts
satisfy the conditions of Theorems \ref{thUimpV} {\em and}
\ref{thVimpU}, i.e.\ $U_{1\infty}$ exists iff $V_{\infty\g}$ in
this case.

The intuition behind Theorem \ref{thVimpU} for binary reward is as
follows: The run-length needs to be small compared to the
quasi-horizon, i.e.\ $o(\ph_k)$, to ensure convergence of
$V_{k\g}$. The condition in Theorem \ref{thVimpU} ensures that the
quasi-horizon $\ph_k=O(k)$ grows at most linearly, hence the
run-length $o(m)$ is a small fraction of the sequence up to $m$.
This ensures that $U_{1m}$ ceases to oscillate.
The formal proof slices $U_{1m}$ in ``curves'' to a weighted
mixture of discounted values $V_{k\g}$. Then $V_{k\g}\to\a$
implies $U_{1m}\to\a$.

\paradot{Proof}
We represent $U_{km}$ as a ($0\leq b_j$-weighted) mixture of
$V_{j\g}$ for $k\leq j\leq m$. The condition $c:=\sup_k{\G_k\over
k\g_k}<\infty$ ensures that the redundant tail $\propto
V_{m+1,\g}$ is ``negligible''. Fix $k$ large enough so that
$|V_{j\g}-\a|<\eps$ $\forall j\geq k$. Then
\bqa\label{prAvDist1}
  \sum_{j=k}^m b_j(\a\mp\eps)
  & \lessgtr & \sum_{j=k}^m b_j U_{1j}
  \;=\; \sum_{j=k}^m {b_j\over\G_j}\sum_{i=j}^m\g_i r_i \;+\;
        \sum_{j=k}^m {b_j\over\G_j}\sum_{i=m+1}^\infty\g_i r_i
\\ \nonumber
  & = & \sum_{i=k}^m\left(\sum_{j=k}^i{b_j\over\G_j}\right)\g_i r_i \;+\;
        \left(\sum_{j=k}^m{b_j\over\G_j}\right)\G_{m+1}V_{m+1,\g}
\eqa
In order for the first term on the r.h.s.\ to be a uniform mixture,
we need
\beq\label{prAvDist2}
  \sum_{j=k}^i{b_j\over\G_j} \;=\; {1\over\g_i}{1\over m-k+1}
  \quad (k\leq i\leq m)
\eeq
Setting $i=k$ and, respectively, subtracting an $i\leadsto i-1$ term we get
\beqn
  {b_k\over\G_k}={1\over\g_k}{1\over m-k+1} \qmbox{and}
  {b_i\over\G_i}=\left({1\over\g_i}-{1\over\g_{i-1}}\right){1\over m-k+1}\geq 0
  \qmbox{for} k<i\leq m
\eeqn
So we can evaluate the $b$-sum in the l.h.s.\ of \req{prAvDist1} to
\bqa\label{prAvDist3} \nonumber
  \sum_{j=k}^m b_j
  & = & {1\over m-k+1}\left[\sum_{j=k+1}^m\left({\G_j\over\g_j}-{\G_j\over\g_{j-1}}\right) + {\G_k\over\g_k}\right]
\\ \nonumber
  & = & {1\over m-k+1}\left[\sum_{j=k}^m\left({\G_j\over\g_j}-{\G_{j+1}\over\g_j}\right) + {\G_{m+1}\over\g_m}\right]
\\
  & = & 1 + {\G_{m+1}\over\g_m(m-k+1)} \;=:\; 1 + c_m
\eqa
where we shifted the sum index in the second equality, and used
$\G_j-\G_{j+1}=\g_j$ in the third equality. Inserting
\req{prAvDist2} and \req{prAvDist3} into \req{prAvDist1} we get
\beqn
  (1+c_m)(\a\mp\eps)
  \;\lessgtr\; \sum_{i=k}^m{1\over m-k+1}r_i + {\G_{m+1}\over \g_m(m-k+1)}V_{m+1,\g}
  \;\lessgtr\; U_{km} + c_m(\a\pm\eps)
\eeqn
Note that the excess $c_m$ over unity in \req{prAvDist3} equals the
coefficient of the tail contribution $V_{m+1,\g}$. The above bound shows that
\beqn
  |U_{km}-\a| \;\leq\; (1+2c_m)\eps \;\leq (1+4c)\eps \qmbox{for} m\geq 2k
\eeqn
Hence $U_{m/2,m}\to\a$, which implies $U_{1m}\to\a$ by Lemma \ref{lemUkm}.
\qed

Theorem \ref{thVimpU} can, for instance, be applied to Example \ref{exUeqV}.
Examples \ref{exVnU} and \ref{exUneqV} demonstrate
that the conditions in Theorem \ref{thVimpU} cannot be dropped.
The following proposition shows more strongly, that the sufficient
condition is actually necessary, i.e.\ cannot be weakened.

\begin{proposition}[\bf\boldmath $V_{\infty\g}\not\Rightarrow U_{1\infty}$]\label{propVnU}\hfill\par
For every monotone $\v\g$ with $\sup_k{\G_k\over k\g_k}=\infty$, there
are $\v r$ for which $V_{\infty\g}$
exists, but not $U_{1\infty}$.
\end{proposition}

\paradot{Proof}
The assumption ensures that there exists a sequence $k_1$, $k_2$, $k_3$, ...
for which
\beqn
  {k_n\g_{k_n}\over\G_{k_n}} \leq {1\over n^2}
  \qmbox{We further choose} k_{n+1}>8k_n
\eeqn
We choose a binary reward sequence with $r_k=1$ iff $k_n\leq k<m_n:=2k_n$.
\bqan
  V_{k_n\g} & = &
  {1\over\G_{k_n}}\sum_{l=n}^\infty \g_{k_l}+...+\g_{2k_l-1}
  \;\leq\; {1\over\G_{k_n}}\sum_{l=n}^\infty k_l\g_{k_l}
\\
  & \leq & \sum_{l=n}^\infty {k_l\g_{k_l}\over\G_{k_l}}
  \;\leq\; \sum_{l=n}^\infty {1\over l^2}
  \;\leq\; {1\over n-1} \;\to\; 0
\eqan
which implies $V_{\infty\g}=0$ by \req{prDiscBin1}. In a sense the
1-runs become asymptotically very sparse. On the other hand,
\bqan
  U_{1,m_n-1} & \geq & \textstyle {1\over m_n}[r_{k_n}+...+r_{m_n-1}]
    \;=\; {1\over m_n}[m_n-k_n] \;=\; \odt \qmbox{but}
\\
  U_{1,k_{n+1}-1} & \leq & \textstyle {1\over k_{n+1}-1}[r_1+...+r_{m_{n-1}}]
    \;\leq\; {1\over 8k_n}[m_n-1] \;\leq\; {1\over 4},
\eqan
hence $U_{1\infty}$ does not exist.
\qed

\section{Average Equals Discounted Value}\label{secAEDV}

Theorem \ref{thUimpV} and \ref{thVimpU} together imply for nearly
all discount types (all in our table) that
$U_{1\infty}=V_{\infty\g}$ if $U_{1\infty}$ and $V_{\infty\g}$
both exist.
But Example \ref{exOsc} shows that there are $\v\g$ for which
simultaneously $\sup_k{\G_k\over k\g_k}=\infty$ {\em and}
$\sup_k{k\g_k\over \G_k}=\infty$, i.e.\ neither Theorem
\ref{thUimpV}, nor Theorem \ref{thVimpU} applies. This happens
for quasi-horizons that grow alternatingly super- and sub-linear.
Luckily, it is easy to also cover this missing case, and we get
the remarkable result that $U_{1\infty}$ equals $V_{\infty\g}$ if
both exist, for {\em any} monotone discount sequence
$\v\g$ and {\em any} reward sequence $\v r$, whatsoever.

\begin{theorem}[Average equals discounted value, \boldmath $U_{1\infty}=V_{\infty\g}$]\label{thUeqV}\hfill\par
Assume monotone $\v\g$ and that $U_{1\infty}$ and $V_{\infty\g}$ exist.
Then $U_{1\infty}=V_{\infty\g}$.
\end{theorem}

\paradot{Proof}
Case 1, $\sup_k{\G_k\over k\g_k}<\infty$:
By assumption, there exists an $\a$ such that $V_{k\g}\to\a$.
Theorem \ref{thVimpU} now implies $U_{1m}\to\a$, hence
$U_{1\infty}=V_{\infty\g}=\a$.

Case 2, $\sup_k{\G_k\over k\g_k}=\infty$: This implies that there
is an infinite subsequence $k_1<k_2<k_3,...$
for which $\G_{k_i}/k_i\g_{k_i}\to\infty$, i.e.
$c_{k_i}:=k_i\g_{k_i}/\G_{k_i}\leq c<\infty$. By assumption, there
exists an $\a$ such that $U_{1m}\to\a$. If we look at the proof of
Theorem \ref{thUimpV}, we see that it still implies
$|V_{k_i\g}-\a|<(1+c_{k_i})\eps\leq(1+2c)\eps$ on this
subsequence. Hence $V_{k_i\g}\to\a$. Since we assumed existence of
the limit $V_{k\g}$ this shows that the limit necessarily equals
$\a$, i.e.\ again $U_{1\infty}=V_{\infty\g}=\a$. \qed

Considering the simplicity of the statement in Theorem
\ref{thUeqV}, the proof based on the proofs of Theorems
\ref{thUimpV} and \ref{thVimpU} is remarkably complex. A simpler
proof, if it exists, probably avoids the separation of the two
(discount) cases.

Example \ref{exUneqV} shows that the monotonicity condition in
Theorem \ref{thUeqV} cannot be dropped.

\section{Discussion}\label{secDisc}

We showed that asymptotically, discounted and average value are
the same, provided both exist. This holds for essentially
arbitrary discount sequences (interesting since geometric discount
leads to agents with bounded horizon) and arbitrary reward
sequences (important since reality is neither ergodic nor MDP).
Further, we exhibited the key role of power discounting with
linearly increasing effective horizon. First, it separates the
cases where existence of $U_{1\infty}$ implies/is-implied-by
existence of $V_{\infty\g}$. Second, it neither requires nor
introduces any artificial time-scale; it results in an increasingly
farsighted agent with horizon proportional to its own age. In
particular, we advocate the use of quadratic discounting
$\g_k=1/k^2$.
All our proofs provide convergence rates, which could be extracted
from them. For simplicity we only stated the asymptotic results.
The main theorems can also be generalized to probabilistic
environments. Monotonicity of $\v\g$ and boundedness of rewards
can possibly be somewhat relaxed. A formal relation between
effective horizon and the introduced quasi-horizon may be
interesting.

\newpage

\begin{small}

\end{small}

\end{document}